\pdfoutput=1
\documentclass[11pt]{article}

\usepackage[preprint]{acl}  

\usepackage{times}
\usepackage{latexsym}
\usepackage[safe]{tipa}

\usepackage[T1]{fontenc}

\usepackage[utf8]{inputenc}

\usepackage{microtype}

\usepackage{graphicx}
\usepackage{booktabs}
\usepackage{multirow}
\usepackage{array}
\usepackage{float}
\usepackage{amsmath}
\usepackage{amssymb}
\usepackage{tikz}
\usetikzlibrary{arrows.meta,positioning}
\newcommand{\hangultext}[2][1.1em]{\raisebox{-0.2ex}{\includegraphics[height=#1]{#2}}}

\title{How Do Language Models Represent and Use Phonological Information for Allomorph Selection?}

\author{Sangwoo Kim \\
  Department of Linguistics \\
  Seoul National University \\
  \texttt{hemy0101@snu.ac.kr}
  \And
  Sangah Lee\thanks{ \quad Corresponding author.} \\
  Department of Linguistics \\
  Seoul National University \\
  \texttt{sanalee@snu.ac.kr}
}

\begin{document}
\maketitle
\begin{abstract}
Language models are trained on tokenized text that obscures the sound structure of words, yet they reliably produce morphemes whose form is phonologically conditioned. It remains unclear whether they rely on item-specific memorization or rule-like generalization and, if the latter, how that generalization is implemented. We therefore ask whether this phonological condition is represented within language models and how it is causally used for allomorph selection. For the English indefinite article \emph{a}/\emph{an}, we show that the phonological condition is encoded along a single linear direction in trigger-token embeddings, that this direction causally drives article selection in token-level \emph{wug} tests, and that, at the article-prediction position, the model forecasts the upcoming trigger token and uses the forecasted trigger's phonological feature to choose the article. We then ask whether this rule-like generalization extends beyond English article selection, both to allomorph selection in other languages and to explicit phonological judgment. Together, these results provide a mechanistic account of phonologically conditioned allomorph selection in language models, and dissociate this generation-time ability from explicit metalinguistic judgments.
\end{abstract}

\section{Introduction}
\label{sec:introduction}

Large language models (LLMs) now extend well beyond standard NLP benchmarks into a growing range of real-world applications \citep{Bommasani2021FoundationModels,zhao2026surveylargelanguagemodels}. Yet because these models are neural networks trained end-to-end on massive amounts of raw text without explicit linguistic supervision \citep{devlin-etal-2019-bert,brown2020language}, the nature of the linguistic knowledge they internalize remains incompletely understood. 

Understanding the linguistic knowledge that LLMs internalize is theoretically important, as they offer a rare view into how a non-human system comes to represent language. It is also practically important, because LLMs are increasingly used in settings where reliable linguistic behavior matters, including language assessment, therapy, and education \citep{yancey-etal-2023-rating, 10.3389/fresc.2025.1600145, davis-etal-2024-prompting}.
These considerations have motivated a growing body of work on what linguistic knowledge LLMs acquire and how they deploy it.

Within this broader effort to understand various linguistic domains, we focus on \emph{morphophonology} --- the domain in which phonology and morphology interact. When humans acquire a spoken language, phonological information is directly available in the acoustic signal. LLMs, in contrast, are trained on text alone, and tokenization further abstracts words into atomic units that do not expose their sound structure, leaving phonology available to them only indirectly \citep{itzhak-levy-2022-models, kaushal-mahowald-2022-tokens}. And yet these models are trained to produce morphemes whose form is phonologically conditioned. Whether their ability to do so reflects item-specific memorization or a rule-like generalization\footnote{Throughout, \emph{rule-like generalization} denotes a property of the model's behavior, not of its implementation. We do not claim that the model encodes or applies a symbolic rule, and our experiments are not designed to settle that question.} remains unclear; if the latter, the mechanism by which the generalization is implemented is unknown. We use phonologically conditioned allomorph selection to ask whether language models represent a phonological condition shared across tokens, rather than tied to particular lexical items, and how they use that representation for allomorph selection. 

Concretely, we focus first on the English indefinite article \emph{a}/\emph{an}, a canonical case of phonologically conditioned allomorphy. We first ask whether the phonological condition governing this choice is encoded in trigger-token embeddings, and show that it is concentrated in a single direction in embedding space (\S\ref{sec:rq1}). We then use token-level wug tests to show that this direction causally controls the model's choice between \emph{a} and \emph{an} (\S\ref{sec:rq2}). To identify the internal mechanism, we use FutureLens \citep{pal-etal-2023-future} and its reverse mapping to show that, at the article-prediction position, the model forecasts the upcoming trigger token and selects the article based on the phonological feature of that forecast (\S\ref{sec:rq3}). Finally, we use the same token-level wug framework to examine the scope of this effect beyond English article selection, finding analogous causal effects in Korean, Turkish, Italian, and French allomorphy but a dissociation from explicit sound-judgment questions (\S\ref{sec:rq4}).

Our results carry a simple memorization-vs-generalization question into a mechanistic account of how the phonological condition for \emph{a}/\emph{an} is represented, used, and routed inside the model, and how far this account extends beyond English article selection. Whereas prior work on phonological knowledge in LLMs has relied largely on metalinguistic benchmarks and output-level evaluation \citep{suvarna-etal-2024-phonologybench, edman-etal-2024-cute}, our analysis contributes by directly inspecting the internal representations and generation mechanisms that produce these behaviors. Our code is available at \url{https://github.com/Koolkumn/allomorph-selection}.

\section{Related Work}

\subsection{Subtoken Information in Language Models}

Language models are usually trained on text, and tokenization makes tokens their basic input units. They therefore lack explicit access to the internal structure of tokens, such as orthographic form, phonological form, or morphological composition \citep{edman-etal-2024-cute,bostrom-durrett-2020-byte,xue-etal-2022-byt5}. Yet the strings they learn to predict come from human languages in which these lower-level properties can be relevant to predicting the next word or morpheme. This mismatch has motivated a growing body of work on how language models handle structure below the token level. Prior work has approached this problem in two broad ways: by modifying models so that they can access such structure more directly, and by asking what pretrained models know about structure below the token level.

One line of work modifies the input or training setup so that models can access structure below the token level, for example by using bytes or characters \citep{xue-etal-2022-byt5,clark-etal-2022-canine}, adding phoneme or IPA information \citep{gale-etal-2023-mixed,nguyen-etal-2023-enhancing}, or incorporating phonological features into morphological models \citep{guriel-etal-2023-morphological}. Studies in these settings show that phonological and morphophonological structure can be made available to neural models, and can sometimes be detected in their internal representations \citep{astrach-pinter-2025-probing,silfverberg-etal-2021-rnn}. They leave open, however, how standard language models handle phonological conditions.

A second line of work analyzes already trained models. Work such as PhonologyBench \citep{suvarna-etal-2024-phonologybench}, CUTE \citep{edman-etal-2024-cute}, EXECUTE \citep{edman-etal-2025-execute}, and \citet{weissweiler-etal-2023-counting} evaluates whether LLMs can use phonological, orthographic, or morphological information in their outputs. However, because these benchmarks evaluate models by asking direct questions about linguistic form, it remains unclear whether success reflects metalinguistic knowledge used to answer the task or information represented and used during ordinary text generation. Thus, these studies do not show how the relevant phonological condition is internally represented or used during prediction.

Most closely related to our work, probing studies show that token embeddings contain substantial character-level information \citep{itzhak-levy-2022-models,kaushal-mahowald-2022-tokens}. Our study builds on this observation by identifying a phonological direction in token embedding space and showing that this direction causally contributes to allomorph selection.

\subsection{Linear Representation Hypothesis}

A growing body of interpretability work suggests that models' internal features are organized linearly in activation space. Under the linear representation hypothesis, a feature can often be read out by projecting activations onto a direction, and in some cases manipulated by intervening along that direction \citep{pmlr-v235-park24c,ICLR2024_0a605985,marks2024the}. This perspective motivates the use of linear probes and steering interventions. However, linearly decodable information may be distributed across a low-dimensional subspace rather than concentrated in a single direction, and decodability alone does not establish causal use \citep{ravfogel-etal-2020-null,elazar-etal-2021-amnesic,pmlr-v162-ravfogel22a}. We therefore treat the dimensionality and causal role of the phonological condition for \emph{a}/\emph{an} selection as empirical questions, first identifying a linear phonological direction and then testing whether the information is concentrated in that single direction and causally used in article prediction.

\section{Preliminaries}

\subsection{Phonologically Conditioned Allomorphy}

Many morphemes have multiple surface realizations, or \emph{allomorphs}, whose distribution is conditioned by phonological, morphological, or lexical context \citep{haspelmath-sims-2010}. When this choice of allomorph is determined by the surrounding sound structure, the phenomenon is known as \emph{phonologically conditioned allomorphy}. Crucially, this process is highly productive: human speakers readily apply the pattern to novel words such as neologisms and borrowings. This behavior suggests that the choice of allomorph is derived from a phonological rule rather than memorized lexically \citep{haspelmath-sims-2010, Berko01081958}. 

To study whether LLMs make such phonologically conditioned choices, we focus on the English indefinite article \emph{a}/\emph{an}, a canonical case of allomorph selection. The relevant contrast is phonological rather than orthographic: \emph{a} appears before consonant-initial words, including orthographic vowel cases such as \emph{university}, whereas \emph{an} appears before vowel-initial words, including orthographic consonant cases such as \emph{hour} (Table~\ref{tab:aan}) \citep{pak-2016-how}. We refer to the following word-start token whose initial sound conditions this choice as the \emph{trigger token}.

\begin{table}[t]
\centering
\small
\begin{tabular}{@{}lll@{}}
\toprule
Allomorph & Initial sound & Examples \\
\midrule
\emph{a}  & consonant & \emph{a banana}; \emph{a university} \textipa{/j-/} \\
\emph{an} & vowel     & \emph{an apple}; \emph{an hour} \textipa{/aU-/} \\
\bottomrule
\end{tabular}
\caption{Allomorphs of the English indefinite article \emph{a}/\emph{an}, conditioned by the initial sound of the trigger token rather than its orthography.}
\label{tab:aan}
\end{table}

\subsection{Notation and Setup}

We analyze decoder-only Transformer language models with $L$ transformer layers and hidden dimension $d$. Let $V$ denote the vocabulary, and let $E \in \mathbb{R}^{|V| \times d}$ be the token embedding matrix. We write $E_i \in \mathbb{R}^d$ for the embedding vector in row $i$ of $E$. At timestep $t$, let $h_t^\ell \in \mathbb{R}^d$ denote the hidden state after transformer layer $\ell$; in particular, $h_t^L$ is the final transformer-layer state before the model's final RMS normalization. The model's next-token distribution at timestep $t$, denoted $P_t \in [0, 1]^{|V|}$, is computed by applying an unembedding map $D(h)$ to $h_t^L$:
\[
\begin{aligned}
P_t &= D(h_t^L),\\
D(h) &= \operatorname{softmax}\left(
W_{\text{out}}
\left(\gamma \odot \frac{h}{\operatorname{RMS}(h)}\right)
\right),
\end{aligned}
\]
where $W_{\text{out}} \in \mathbb{R}^{|V| \times d}$ is the output projection matrix, $\gamma \in \mathbb{R}^d$ is the effective scaling vector of the final RMS normalization, and $\odot$ denotes the element-wise product.

For a token $x \in V$, $P_t(x)$ denotes the probability assigned to $x$ by this distribution. When the timestep is clear from context, we write this probability simply as $P(x)$. For weight-tied models, the output projection is the same matrix as the input embedding matrix; in this case, we write $W_{\text{out}}=E$.

For English \emph{a}/\emph{an}, our main analysis proceeds in the three stages previewed above. We first identify a phonological direction in trigger-token embeddings (\S\ref{sec:rq1}). We then test whether that direction is causally used for article prediction (\S\ref{sec:rq2}). Third, we examine how the relevant information is routed through the model (\S\ref{sec:rq3}). We evaluate these three stages across three instruction-tuned decoder-only language models: Llama-3.2-3B-Instruct \citep{grattafiori2024llama3herdmodels}, Qwen2.5-3B-Instruct \citep{qwen2.5}, and Gemma-3-1b-it \citep{gemma_2025}. These models all use tied input and output embeddings, and their tokenizers each represent the article forms ``\texttt{ a}'' and ``\texttt{ an}'' as single tokens. In addition, we ask whether the same embedding-based account extends beyond English \emph{a}/\emph{an} (\S\ref{sec:rq4}). Artifact details, licenses, and terms of use are reported in Appendix~\ref{app:responsible-research}.

\section{A Phonological Direction in Token Embedding Space}
\label{sec:rq1}

If models select the \emph{a}/\emph{an} allomorphs through a rule-like generalization, analogous to human speakers, then the phonological feature that conditions this choice should be encoded somewhere in the model. Prior work shows that token embeddings encode character-level and orthographic information about tokens \citep{itzhak-levy-2022-models, kaushal-mahowald-2022-tokens}; following this line of work, we ask whether the phonological feature that determines \emph{a}/\emph{an} selection is represented in token embedding space, and how that representation is organized.

\subsection{Identifying a Phonological Direction}

We first identify a phonological direction by training a linear classifier to distinguish vowel-initial from consonant-initial trigger tokens using their token embeddings. Because English spelling usually predicts whether a word begins with a vowel or consonant sound, a classifier could achieve high accuracy by tracking spelling alone. We therefore train the classifier on tokens whose orthographic and phonological labels agree, and evaluate it on exception tokens whose labels conflict.

Concretely, we assign each token that functions as a trigger token in the English Wikipedia corpus two labels.\footnote{We use the English Wikipedia snapshot \texttt{20231101.en} distributed in the Hugging Face \texttt{wikimedia/wikipedia} dataset (\url{https://huggingface.co/datasets/wikimedia/wikipedia}).} The phonological label indicates whether the token behaves as vowel-initial or consonant-initial for \emph{a}/\emph{an} selection, and the orthographic label is a simple first-letter heuristic, defined by whether the written form begins with one of \emph{a, e, i, o, u}. Tokens whose two labels agree are used for training, while tokens whose labels conflict are held out as an exception test set (Table~\ref{tab:rq1-split}). Thus, the classifier is trained on ordinary positive cases such as \emph{apple} and \emph{ordinary}, and ordinary negative cases such as \emph{banana} and \emph{city}, but evaluated on exception positives such as \emph{hour}, \emph{honor}, and \emph{heir}, and exception negatives such as \emph{one}, \emph{unique}, and \emph{university}, where spelling alone predicts the wrong allomorph.

\begin{table}[t]
\centering
\small
\setlength{\tabcolsep}{2pt}
\begin{tabular}{@{}ccc@{}}
\toprule
\multirow{2}{*}{Orthography} & \multicolumn{2}{c@{}}{Phonology} \\
\cmidrule(l){2-3}
 & consonant-start & vowel-start \\
\midrule
consonant-start & \shortstack{Train\\\emph{banana}, \emph{city}} & \shortstack{Test\\\emph{hour}, \emph{honor}, \emph{heir}} \\
vowel-start & \shortstack{Test\\\emph{one}, \emph{unique}, \emph{university}} & \shortstack{Train\\\emph{apple}, \emph{ordinary}} \\
\bottomrule
\end{tabular}
\caption{Train/test split for fitting the phonological direction. Training tokens are cases where orthography and corpus-derived phonological label agree; test tokens are exceptions where orthography predicts the wrong allomorph.}
\label{tab:rq1-split}
\end{table}

For each retained token, we use its embedding as the input to a binary logistic regression classifier. We normalize the classifier's weight vector and denote the resulting direction by $\mathbf{d}_1 \in \mathbb{R}^{d}$. If $\mathbf{d}_1$ captures a genuinely phonological distinction, it should classify the exception tokens correctly, not merely perform well on ordinary tokens where spelling and sound agree.

Across all three models, the learned direction generalizes strongly to the held-out exception tokens (Table~\ref{tab:rq1-main}). Cross-validation accuracy exceeds 99.8\% in every case, and exception-set accuracy remains high even though a first-letter heuristic would predict the wrong article. This indicates that trigger token embeddings encode a phonological vowel/consonant distinction, not merely an orthographic one.

\begin{table}[t]
\centering
\small
\resizebox{\linewidth}{!}{%
\begin{tabular}{lrrcl}
\toprule
Model & Train $n$ & CV acc. & Test correct & Misses \\
\midrule
Llama & 8,934 & 99.89 & 41/41 & -- \\
Qwen & 8,915 & 99.85 & 39/41 & \emph{heir}, \emph{honors} \\
Gemma & 9,873 & 99.93 & 49/50 & \emph{hei} \\
\bottomrule
\end{tabular}
}
\caption{Phonological-direction fitting results. Train $n$ counts retained trigger-token vocabulary items and CV accuracy is a percentage; the test column reports correctly classified tokens out of the full exception set. The test set consists only of orthography--phonology exception tokens.}
\label{tab:rq1-main}
\end{table}

\subsection{Testing the Dimensionality of the Representation}

Prior work on probing and concept erasure has shown that a linearly decodable property need not be confined to a single direction: removing one classifier direction can leave residual signal that another linear classifier can recover \citep{ravfogel-etal-2020-null, pmlr-v162-ravfogel22a}. Following Iterative Null-space Projection \citep{ravfogel-etal-2020-null}, we therefore ask whether the phonological condition for \emph{a}/\emph{an} is concentrated in $\mathbf{d}_1$, or instead distributed across a larger linear subspace.

To test this, we remove $\mathbf{d}_1$ from all token embeddings and train a second classifier on the residual embeddings, keeping the train/test split identical to the first:
\[
E' = E - (E \mathbf{d}_1)\mathbf{d}_1^\top .
\]
If the \emph{a}/\emph{an} condition is distributed across multiple independent directions, the second classifier should remain predictive; if its performance collapses, the information is concentrated in a single dominant direction. 

Across all three models, the second classifier fails to exceed chance on cross-validation once $\mathbf{d}_1$ is projected out (Table~\ref{tab:rq1-iterative}),\footnote{The training tokens are class-imbalanced, so the chance level is not 50\% (Appendix~\ref{app:rq1-details}).} and a non-linear MLP probe on the residual embeddings also stays near chance. The phonological feature is therefore encoded in a one-dimensional subspace of the embedding space, and we refer to this direction $\mathbf{d}_1$ simply as $\mathbf{d}$ in the remainder of the paper. The MLP probe, the stability of $\mathbf{d}$, and other experimental details are reported in Appendix~\ref{app:rq1-details}.

\begin{table}[t]
\centering
\small
\resizebox{\linewidth}{!}{%
\begin{tabular}{lrrrr}
\toprule
\multirow{2}{*}{Model} & \multicolumn{2}{c}{$\mathbf{d}_1$} & \multicolumn{2}{c}{Residual after removing $\mathbf{d}_1$} \\
\cmidrule(lr){2-3}\cmidrule(l){4-5}
 & CV acc. & Test acc. & CV acc. & Test acc. \\
\midrule
Llama & 99.89 & 100.0 & 40.82 & 73.2 \\
Qwen & 99.85 & 95.1 & 38.31 & 61.0 \\
Gemma & 99.93 & 98.0 & 51.74 & 42.0 \\
\bottomrule
\end{tabular}
}
\caption{Dimensionality test for the phonological embedding direction. All entries are accuracies in percent. The first two columns evaluate the original classifier direction $\mathbf{d}_1$; the last two evaluate a new residual classifier trained after projecting $\mathbf{d}_1$ out of all embeddings. Test accuracy is measured on the same held-out exception set as Table~\ref{tab:rq1-main}.}
\label{tab:rq1-iterative}
\end{table} 

\section{Causal Use in Token-Level Wug Tests}
\label{sec:rq2}

Section~\ref{sec:rq1} showed that the phonological feature relevant to \emph{a}/\emph{an} selection is encoded along a single direction $\mathbf{d}$ in token embedding space, but this result does not by itself show that the model uses this direction when choosing between the two allomorphs. If a model's choice for a previously unseen trigger token tracks that token's component along $\mathbf{d}$, this would provide causal evidence that the model uses the phonological feature productively, rather than relying on memorized article preferences for familiar lexical items. This is the logic of the classical wug test \citep{Berko01081958}, in which speakers inflect nonce words such as \emph{wug}. English speakers choose the plural allomorph for such items from the stem's final segment alone, showing that the pattern is applied by rule rather than retrieved from memory. We design a token-level version of this test, using nonce trigger tokens, the model's native lexical units, whose embeddings vary only in their component along $\mathbf{d}$.

\subsection{Token-Level Wug Test}

We construct each nonce trigger token by sampling a base embedding from the empirical distribution of ordinary trigger-token embeddings, projecting out its $\mathbf{d}$ component, and adding back a chosen amount of $\mathbf{d}$. This construction keeps each nonce embedding statistically close to real trigger-token embeddings, while ensuring that controlled variants of the same nonce token differ only along $\mathbf{d}$. We validate the sampled nonce bases in Appendix~\ref{app:rq2-details}.

Concretely, for each model we let $\mu \in \mathbb{R}^{d}$ and $\Sigma \in \mathbb{R}^{d \times d}$ be the mean and covariance of the ordinary trigger-token embeddings used to identify $\mathbf{d}$, and sample 500 base embeddings $z \in \mathbb{R}^{d}$ with $z \sim \mathcal{N}(\mu,\Sigma)$. We remove the $\mathbf{d}$ component of each,
\[
\tilde{z} = z - (z^\top \mathbf{d})\mathbf{d},
\]
and add back a chosen amount of the phonological direction to obtain controlled variants of the nonce token,
\[
z(\alpha) = \tilde{z} + \alpha \mathbf{d}.
\]
Let $\bar\alpha_c$ and $\bar\alpha_v$ be the mean values of $E_i^\top \mathbf{d}$ over the consonant-start and vowel-start training tokens from Section~\ref{sec:rq1}. We sweep $\alpha$ over 30 values symmetrically around the midpoint $(\bar\alpha_c+\bar\alpha_v)/2$, extending half a gap beyond $\bar\alpha_c$ and $\bar\alpha_v$.

We insert each $z(\alpha)$ into the embedding row of a reserved nonce token and place that token as the trigger in a Chinese-to-English translation prompt. The Chinese source sentence does not specify which English article form should be used, so the model should choose between \emph{a} and \emph{an} when producing the English translation. We use the template in Table~\ref{tab:rq2-template}, and measure the next-token probabilities assigned to \emph{a} and \emph{an} after the English prefix.

\begin{table}[t]
\centering
\small
\begin{tabular}{@{}m{0.68\columnwidth}m{0.20\columnwidth}@{}}
\toprule
Source prompt & Prefix \\
\midrule
\includegraphics[width=0.68\columnwidth]{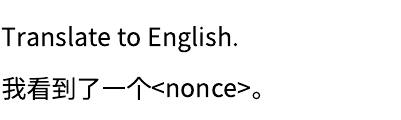} & I saw \\
\bottomrule
\end{tabular}
\caption{Translation template for the token-level wug test. The \texttt{<nonce>} placeholder is replaced by the reserved nonce token for each model.}
\label{tab:rq2-template}
\end{table}

\begin{figure}[t]
\centering
\includegraphics[width=\columnwidth]{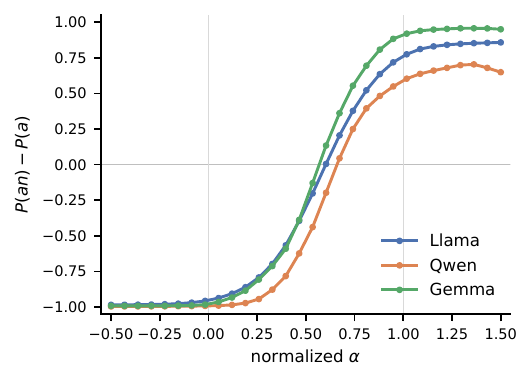}
\caption{Effect of varying the $\mathbf{d}$ component of nonce trigger tokens in the token-level wug test. Each line shows one model; the plotted value is $P(an)-P(a)$, averaged over 500 nonce bases in the Chinese-to-English translation template. The $x$-axis normalizes $\alpha$ so that 0 is the consonant-start class mean and 1 is the vowel-start class mean.}
\label{fig:rq2-wug-translation}
\end{figure}

The nonce-token sweep strongly shifts article prediction in all three models (Figure~\ref{fig:rq2-wug-translation}). Near the consonant-start anchor, the score $P(an)-P(a)$ is strongly negative, reflecting a preference for \emph{a}; near the vowel-start anchor, it becomes positive, reflecting a preference for \emph{an}, though not as strong as the preference for \emph{a}.

Since the nonce token is held fixed except for its embedding component along $\mathbf{d}$, this result shows that the phonological direction is not merely decodable from token embeddings, but causally affects the model's allomorph choice. Crucially, the nonce token never appears in the model's training data, so the model cannot have memorized any article preference for it; the causal effect of $\mathbf{d}$ therefore reflects a productive phonological generalization rather than lexical recall. The sweep also reveals an asymmetry: $|P(an)-P(a)|$ is consistently larger at the consonant-start anchor than at the vowel-start anchor, indicating a stronger preference for \emph{a} than for \emph{an}. This asymmetry is consistent with the base-rate skew of English, in which consonant-initial words --- and hence \emph{a} --- vastly outnumber vowel-initial words and \emph{an}; the model's prior appears to inherit this skew. Experimental details for this section are reported in Appendix~\ref{app:rq2-details}.

\section{Forecasting the Trigger During Article Prediction}
\label{sec:rq3}

Section~\ref{sec:rq2} showed that the trigger token's component along the phonological direction $\mathbf{d}$ causally affects the model's choice between \emph{a} and \emph{an}. We now ask how this effect is implemented inside the model. For human speakers, selecting between \emph{a} and \emph{an} is naturally conditioned on the following word: speakers choose the article by considering the initial sound of the upcoming trigger word. Decoder-only transformers face a different situation. Because they generate text autoregressively, the model cannot explicitly access the trigger token when predicting \emph{a}/\emph{an}. We hypothesize that the model, like human speakers, anticipates the trigger token when selecting \emph{a}/\emph{an} and chooses the article based on the phonological feature of that anticipated trigger.

\subsection{Forecasting the Trigger from the Prediction Position}

\begin{figure}[t]
\centering
\resizebox{\columnwidth}{!}{%
\begin{tikzpicture}[
    >=Latex,
    state/.style={draw=gray!62, rounded corners=1.5pt, minimum width=.50cm,
        minimum height=.42cm, inner sep=0pt, line width=.42pt, fill=white},
    source/.style={state, draw=blue!55!black, line width=.72pt,
        minimum width=.78cm, font=\small\bfseries},
    predict/.style={state, draw=blue!65!black, fill=blue!7,
        line width=.85pt, minimum width=.78cm, font=\small\bfseries},
    future/.style={predict, minimum width=1.02cm},
    map/.style={draw=blue!55!black, fill=blue!6, rounded corners=2pt,
        inner sep=2pt, font=\small\bfseries},
    comp/.style={->, draw=gray!33, line width=.42pt},
    shaft/.style={draw=gray!27, line width=.42pt},
    mainarrow/.style={->, draw=blue!60!black, line width=.95pt},
    decode/.style={->, draw=black!70, line width=.82pt},
    token/.style={font=\scriptsize\sffamily\itshape, anchor=base}
]
\def\xc{0}
\def\xn{1.9}
\def\xa{3.8}
\def\xt{5.7}
\def\ye{0}
\def\ydown{.75}
\def\yl{1.5}
\def\yup{2.25}
\def\yL{3.0}
\def\ydecode{3.54}
\def\ytoken{3.73}

\foreach \y in {\ye,\yl,\yL} {
    \draw[gray!14, line width=.45pt] (-.58,\y) -- (6.30,\y);
}
\foreach \x in {\xc,\xn,\xa,\xt} {
    \draw[shaft] (\x,.30) -- (\x,.54);
    \node[text=gray!43, font=\scriptsize] at (\x,\ydown) {$\vdots$};
    \draw[comp] (\x,.96) -- (\x,1.16);
    \draw[shaft] (\x,1.84) -- (\x,2.04);
    \node[text=gray!43, font=\scriptsize] at (\x,\yup) {$\vdots$};
    \draw[comp] (\x,2.46) -- (\x,2.66);
}

\node[black, font=\small] at (-1.25,\yL) {Layer $L$};
\node[text=gray!55, font=\small] at (-1.25,\yup) {$\vdots$};
\node[black, font=\small] at (-1.25,\yl) {Layer $\ell$};
\node[text=gray!55, font=\small] at (-1.25,\ydown) {$\vdots$};
\node[black, font=\small] at (-1.25,\ye) {Embed.};

\node[state] (cL) at (\xc,\yL) {};
\node[state, minimum width=.78cm, font=\small] (nL) at (\xn,\yL) {$h_t^L$};
\node[future] (aL) at (\xa,\yL) {$h_{t+1}^{L}$};
\node[state] (tL) at (\xt,\yL) {};

\node[state] (cl) at (\xc,\yl) {};
\node[predict] (nl) at (\xn,\yl) {$h_t^\ell$};
\node[state] (al) at (\xa,\yl) {};
\node[state] (tl) at (\xt,\yl) {};

\node[state] (ce) at (\xc,\ye) {};
\node[state] (ne) at (\xn,\ye) {};
\node[state] (ae) at (\xa,\ye) {};
\node[state] (te) at (\xt,\ye) {};

\draw[mainarrow] (nl.north east) -- node[map, pos=.55, above left=1pt] {$F_\ell$} (aL.south west);
\draw[decode] (nL.north) -- (\xn,\ydecode);
\draw[decode] (aL.north) -- (\xa,\ydecode);
\node[token] at (\xn,\ytoken) {an};
\node[token] at (\xa,\ytoken) {apple};

\draw[gray!45, line width=.45pt] (-.35,-.45) -- (6.05,-.45);
\foreach \x in {\xc,\xn,\xa,\xt} {
    \draw[gray!45, line width=.45pt] (\x,-.51) -- (\x,-.39);
}
\node[font=\scriptsize, align=center] at (\xc,-.78) {$\cdots$};
\node[font=\scriptsize, align=center] at (\xn,-.78) {before article\\$t$};
\node[font=\scriptsize, align=center] at (\xa,-.78) {article\\$t+1$};
\node[font=\scriptsize, align=center] at (\xt,-.78) {trigger\\$t+2$};
\end{tikzpicture}%
}
\caption{Schematic of the FutureLens map. From the article prediction position $t$, $F_\ell$ maps an intermediate state $h_t^\ell$ to a forecast of the article-position final state $h_{t+1}^L$, whose decoding predicts the trigger token.}
\label{fig:rq3-futurelens-schematic}
\end{figure}
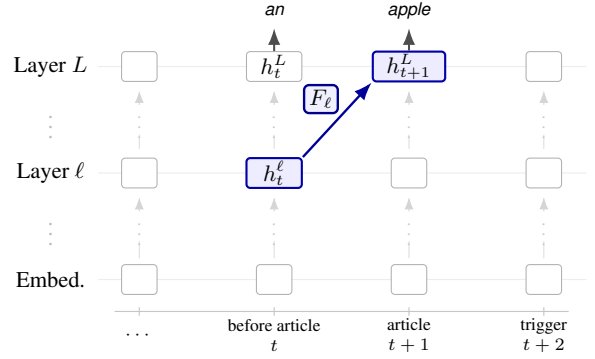

We first ask whether the hidden state at the article-prediction position already contains a forecast of the upcoming trigger token. We use FutureLens \citep{pal-etal-2023-future}, a framework for testing whether a single hidden state anticipates not only the next token but also the one after it. Let $t$ denote the article-prediction position and $t+1$ the article position: the model predicts \emph{a}/\emph{an} from $h_t^L$, and predicts the following trigger token from $h_{t+1}^L$. For each example, a single forward pass gives us both the intermediate state $h_t^\ell$ at every layer $\ell$ and the final state $h_{t+1}^L$; we cache these as source--target pairs and, for each layer, fit a FutureLens map
\[
F_\ell(h_t^\ell) = W_\ell h_t^\ell + \mathbf{b}_\ell
\]
from the source to the target (Figure~\ref{fig:rq3-futurelens-schematic}).

Using the unembedding map $D$ defined above, we train $F_\ell$ over training examples $\mathcal{D}$ to make the decoded distribution of the predicted future state match $P_{t+1}$, the model's output distribution at the article position:
\[
\mathcal{L}_\ell = \mathbb{E}_\mathcal{D}\bigl[\mathrm{KL}\bigl(P_{t+1} \,\|\, D(F_\ell(h_t^\ell))\bigr)\bigr].
\]
Each example consists of two consecutive Wikipedia sentences: the first provides left context, and the second contains a target occurrence of \emph{a} or \emph{an}. Following \citet{pal-etal-2023-future}, we apply a top-1 consistency filter: we retain only occurrences for which decoding from $h_t^L$ predicts the observed article, and decoding from $h_{t+1}^L$ predicts the following word-start trigger token. We constrain $W_\ell$ to be orthogonal so that the linear component of the FutureLens map is invertible in closed form, with $W_\ell^{-1} = W_\ell^\top$.\footnote{Appendix~\ref{app:rq3-details} compares ordinary and orthogonal FutureLens maps and shows that the constraint preserves the late-layer forecasting accuracy on which our analysis relies.} At evaluation, we ask whether the top-$k$ tokens decoded from $F_\ell(h_t^\ell)$ include the ground-truth trigger token at $t+2$. For the random-token control, we keep the context construction and top-1 filtering criteria the same, but replace the \emph{a}/\emph{an} target position with a randomly selected token in the second sentence.

\begin{figure}[t]
\centering
\includegraphics[width=\columnwidth]{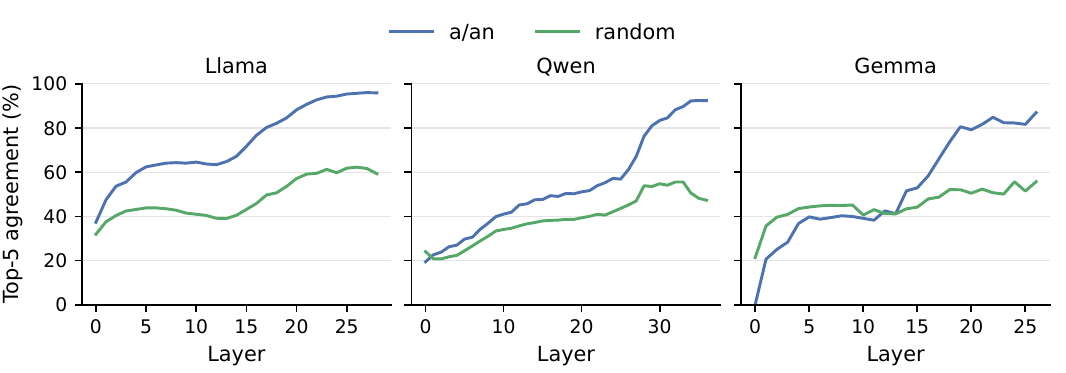}
\caption{FutureLens top-5 agreement across layers. The probe maps the hidden state before a target token to the final state at the target-token position, whose decoding predicts the following token. The \emph{a}/\emph{an} condition is compared with random-token controls.}
\label{fig:rq3-futurelens}
\end{figure}

The FutureLens probe is much more predictive for \emph{a}/\emph{an} examples than for matched random-token controls in late layers (Figure~\ref{fig:rq3-futurelens}), with peak top-5 agreement around 90\% across models. This indicates that, in late layers at the position where the model chooses between \emph{a} and \emph{an}, the hidden state contains a future-state forecast that predicts the following trigger token.

\subsection{Steering the Forecasted Phonological Feature}

We next ask whether the phonological component of this forecast is causally used for article prediction. We do so by mapping the phonological direction $\mathbf{d}$ back to the position where the article is predicted and testing whether intervention along this direction changes the model's \emph{a}/\emph{an} prediction.

Ignoring the final normalization and the bias term, the FutureLens matrix $W_\ell$ maps a vector in the hidden state space at the article-prediction position $t$ to the unembedding space at position $t+1$, where the model predicts the trigger token. Because we constrain $W_\ell$ to be orthogonal, its transpose $W_\ell^\top$ can be viewed as the inverse map from this unembedding space back to the hidden state space at $t$. With weight tying, the phonological direction $\mathbf{d}$ learned in embedding space can be used directly as the corresponding direction in unembedding space. We therefore map $\mathbf{d}$ back through $W_\ell^\top$ and define a phonological direction in the hidden state space at $t$:
\[
\mathbf{s}_\ell = \frac{W_\ell^\top \mathbf{d}}{\|W_\ell^\top \mathbf{d}\|}.
\]

For each layer, we compute the mean projection of training examples with \emph{a} and \emph{an} onto $\mathbf{s}_\ell$, denoted $\mu_a^{(\ell)}$ and $\mu_{an}^{(\ell)}$. At test time, we replace the component of $h_t^\ell$ along $\mathbf{s}_\ell$ with one of these class-mean values:
\[
\tilde{h}_t^\ell
= h_t^\ell
- ((h_t^\ell)^\top\mathbf{s}_\ell)\mathbf{s}_\ell
+ \alpha \mathbf{s}_\ell,
\]
where $\alpha \in \{\mu_a^{(\ell)}, \mu_{an}^{(\ell)}\}$. We evaluate this intervention as a counterfactual top-1 test. From the held-out strict test set, we select 1,000 examples where the unmodified model predicts \emph{a} as its top-1 article and 1,000 where it predicts \emph{an}. We then set $\alpha=\mu_{an}^{(\ell)}$ for the baseline-\emph{a} examples and $\alpha=\mu_a^{(\ell)}$ for the baseline-\emph{an} examples, continue the forward pass, and count how often the top-1 article flips to the opposite form.

\begin{figure}[t]
\centering
\includegraphics[width=\columnwidth]{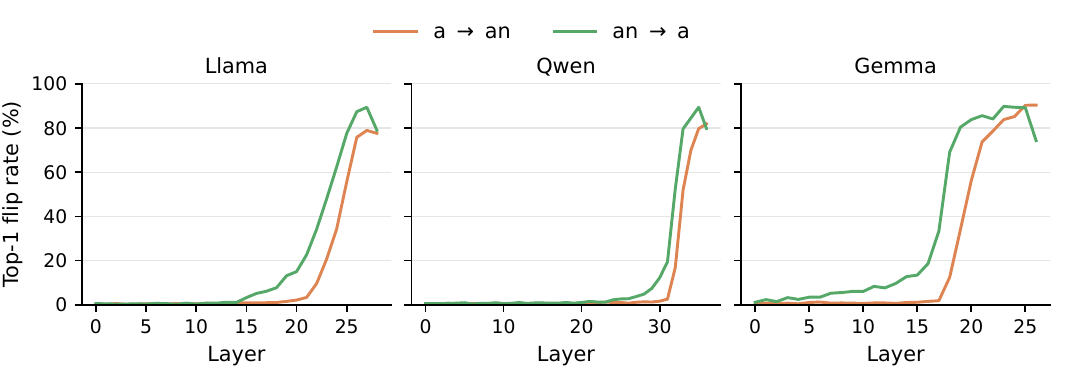}
\caption{Counterfactual steering along the FutureLens-transposed phonological direction. For each layer, baseline-\emph{a} examples are steered to the \emph{an} class mean and baseline-\emph{an} examples are steered to the \emph{a} class mean. The plot reports top-1 flip rates on held-out examples.}
\label{fig:rq3-steering}
\end{figure}

Steering this single direction flips a large majority of the model's article choices in late layers (Figure~\ref{fig:rq3-steering}). Across all three models, peak flip rates reach roughly 84--90\%, indicating that the FutureLens-transposed phonological direction has a strong causal effect on the article decision at the prediction position. Experimental details for this section are reported in Appendix~\ref{app:rq3-details}.

Together, Sections~\ref{sec:rq1}--\ref{sec:rq3} show that the phonological feature required for \emph{a}/\emph{an} selection is represented as a single direction $\mathbf{d}$ in trigger-token embedding space, that this direction is causally used in article selection, and that the model selects between \emph{a} and \emph{an} by internally forecasting the upcoming trigger token and using its phonological feature to determine the appropriate allomorph.

\section{Generalization Beyond English Article Selection}
\label{sec:rq4}

The preceding sections showed that English \emph{a}/\emph{an} selection is supported by a causal phonological direction in trigger-token embeddings. We now ask whether analogous causal effects appear in allomorph selection across other models and languages, and whether the English \emph{a}/\emph{an} direction affects the model's answers to explicit questions about the trigger token's sound. In addition, we verify that the forecasting mechanism of Section~\ref{sec:rq3} also holds in a large model without weight tying; this experiment and its results are reported in Appendix~\ref{app:70b}.

\subsection{Cross-Linguistic Allomorph Selection}

We select four allomorph systems that vary in morphological type and language family: the Korean object particle \emph{-eul}/\emph{-reul}, the Turkish locative suffix \emph{-de}/\emph{-te} vs. \emph{-da}/\emph{-ta}, the Italian indefinite article \emph{un}/\emph{uno}, and the French definite article \emph{le}/\emph{la} vs. \emph{l'}. In Korean and Turkish the trigger precedes the allomorph, whereas in Italian and French it follows. In each case the conditioning feature is binary, but the set of surface forms is not always binary: Turkish alternates four locative forms and French three definite articles.

\begin{figure}[t]
\centering
\includegraphics[width=\columnwidth]{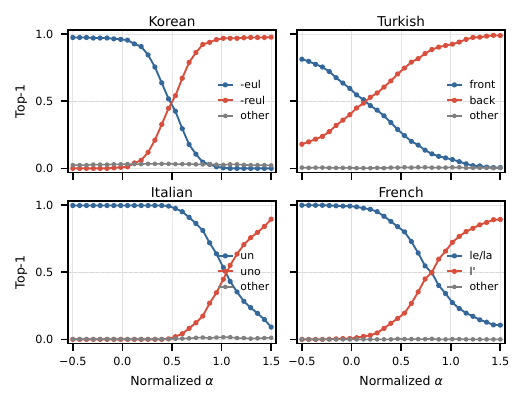}
\caption{Cross-linguistic token-level wug interventions. Each panel plots top-1 rates for the two allomorph families and for other outputs. The $x$-axis normalizes $\alpha$ so that 0 and 1 are the two class means along the learned direction.}
\label{fig:rq4-multilingual-wug}
\end{figure}

Following the same token-level wug protocol as in Section~\ref{sec:rq2}, for each language we learn a linear direction in the target model's embedding space from the corresponding Wikipedia snapshot,\footnote{We use the \texttt{20231101.fr}, \texttt{20231101.ko}, \texttt{20231101.tr}, and \texttt{20231101.it} snapshots of the same \texttt{wikimedia/wikipedia} dataset.} sample nonce embeddings from the empirical distribution of the corresponding trigger tokens, and vary only the nonce token's component along that direction. Table~\ref{tab:rq4-languages} lists the model and the conditioning feature for each language.

\begin{table}[t]
\centering
\scriptsize
\setlength{\tabcolsep}{3pt}
\begin{tabular}{@{}l>{\raggedright\arraybackslash}p{0.40\columnwidth}>{\raggedright\arraybackslash}p{0.30\columnwidth}@{}}
\toprule
Language & Model & Conditioning feature \\
\midrule
Korean & EXAONE-4.0-1.2B \citep{exaone-4.0} & presence of a final consonant \\
\addlinespace
Turkish & Trendyol-LLM-8B-T1 \citep{trendyolLLM8BT1} & backness of the preceding vowel \\
\addlinespace
Italian & Minerva-7B-instruct-v1.0 \citep{orlando-etal-2024-minerva} & phonological shape of the onset \\
\addlinespace
French & Lucie-7B-Instruct-v1.1 \citep{openllm2025lucie} & vowel-initiality of the trigger \\
\bottomrule
\end{tabular}
\caption{Cross-linguistic allomorphy cases used in the token-level wug intervention. The last column gives the phonological feature that conditions the choice in each language.}
\label{tab:rq4-languages}
\end{table}

Figure~\ref{fig:rq4-multilingual-wug} shows language-specific skews in baseline allomorph preference, but in all four cases the learned direction produces a clear top-1 shift toward the corresponding allomorph. This suggests that the causal embedding-direction effect identified for English is not specific to English, a single model, or a single allomorph system. We further apply the FutureLens analysis of Section~\ref{sec:rq3} to Italian and French, where the trigger follows the allomorph, and find that, as in English, the model internally forecasts the upcoming trigger token and uses the forecasted trigger's phonological feature to select the allomorph; this experiment and its results are reported in Appendix~\ref{app:rq4-futurelens}.

\subsection{Dissociation from Explicit Sound Judgment}

For human speakers, allomorph selection and explicit judgments about sound draw on a single phonological representation: the sound form of a word is available both to the grammatical process that selects an allomorph and to conscious reflection about pronunciation. A language model has no such guarantee. It is trained on text alone, with no phonetic transcription or other explicit signal about pronunciation, and the two behaviors are supported by different evidence in that text: allomorph selection can be learned directly from which article precedes each token, whereas statements about sounds come from metalinguistic descriptions of pronunciation. We therefore ask whether the English phonological direction that causally drives \emph{a}/\emph{an} selection is also recruited when the model is explicitly asked about a nonce token's sound. 

We use the same nonce trigger tokens and the same $\alpha$ sweep as in Section~\ref{sec:rq2}, but replace the translation prompt with yes/no questions asking whether the nonce noun begins with a consonant or vowel sound (Table~\ref{tab:rq4-sound-template}).

\begin{table}[t]
\centering
\small
\begin{tabular}{@{}>{\raggedright\arraybackslash}m{0.24\columnwidth}>{\raggedright\arraybackslash}m{0.70\columnwidth}@{}}
\toprule
Question & Prompt \\
\midrule
Sound judgment & Answer yes or no only.\newline Consider the English phrase ``\emph{the}\texttt{<nonce>}''.\newline Does the noun begin with a \{\emph{consonant}, \emph{vowel}\} sound in English pronunciation? \\
\bottomrule
\end{tabular}
\caption{Explicit sound-judgment prompts. The \texttt{<nonce>} placeholder is replaced by the reserved nonce token for each model (with no space after \emph{the}), and \{\emph{consonant}, \emph{vowel}\} indicates two prompt variants used separately.}
\label{tab:rq4-sound-template}
\end{table}

\begin{figure}[t]
\centering
\includegraphics[width=\columnwidth]{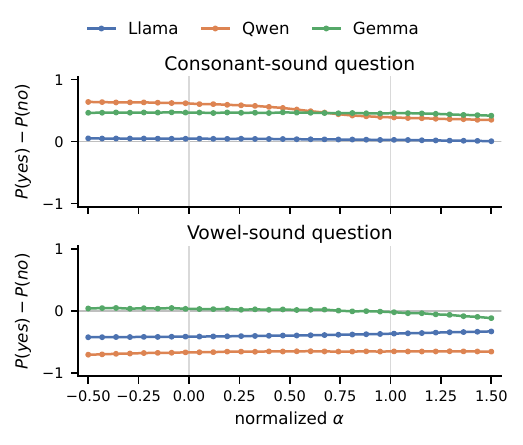}
\caption{Effect of varying the $\mathbf{d}$ component of nonce trigger tokens in explicit sound-judgment prompts. The upper panel asks whether the nonce noun begins with a consonant sound, and the lower panel asks whether it begins with a vowel sound. The plotted value is $P(yes)-P(no)$, averaged over 500 nonce embeddings.}
\label{fig:rq4-sound-judgment}
\end{figure}

Unlike article prediction, the explicit sound judgments show much weaker and less consistent effects of $\mathbf{d}$, with no corresponding monotonic shift toward the expected answer (Figure~\ref{fig:rq4-sound-judgment}). Thus, the direction that LLMs use for allomorph selection is not reliably exposed through explicit sound judgments, supporting a dissociation between internal phonological features for generation and metalinguistic reports about sound.

\section{Conclusion}

Starting from the question of how language models choose between phonologically conditioned allomorphs, we showed that the phonological feature distinguishing English \emph{a} and \emph{an} is encoded as a one-dimensional direction in token embeddings, that this direction causally shapes article generation, and that the model deploys it through an internal forecast of the upcoming trigger token. We then asked whether this rule-like generalization extends beyond English \emph{a}/\emph{an}, both to allomorphy in other languages and to a different kind of task.

We show that language models acquire a generalization for phonologically conditioned allomorphy, a phenomenon widely attested across languages, and that this generalization appears across multiple model families and languages. In addition, the analysis methods introduced to reveal these generalizations are themselves extensible. The token-level wug test introduced here offers a general protocol for probing phonological and other latent linguistic generalizations in LLMs through nonce-token manipulation, and extends naturally beyond a single linear direction to higher-dimensional subspaces or other constructions of nonce embeddings. Similarly, our inverse FutureLens mapping and its use for steering extend FutureLens beyond next-token prediction into a tool for causal intervention: any feature decodable from a hidden state's future forecast can be mapped back to the present position to test its causal effect on the model's current decision. At the same time, the connection and dissociation between the phonological information used for generation and the phonological knowledge accessed when the model is asked directly about sound deserve a more detailed analysis in future work, and should also inform the design of future benchmarks.

\section*{Limitations}

Our study has several limitations. First, our analysis is restricted to decoder-only language models with conventional tokenizers, and does not cover models that depart from standard tokenization, such as BLT-style architectures \citep{pagnoni-etal-2025-byte}. We also do not analyze multimodal LLMs that process speech together with text; extending the present approach to such models would be an interesting direction for future work.

Second, to obtain phonological labels at scale, we derive them from empirical corpus patterns rather than from observed pronunciations. Although we apply strict filtering, we cannot rule out noise in these corpus-derived labels, and some retained items may therefore provide imperfect proxies for phonological class.

Finally, the mechanism described in Section~\ref{sec:rq3} is inferred through an auxiliary FutureLens probe and is therefore only a limited account of the model's internal computation. Our analysis shows that a forecasted phonological feature can be mapped back to the article-prediction position and causally steered, but it does not identify the exact computations by which the model implements this mechanism. A more circuit-level account of these computations remains an important direction for future work.


\bibliography{custom}

\appendix

\section{Experimental Details for Section~\ref{sec:rq1}}
\label{app:rq1-details}

\paragraph{Corpus scan.}
Trigger-token counts were collected from the English Wikipedia snapshot \texttt{20231101.en} in the Hugging Face \texttt{wikimedia/wikipedia} dataset (Table~\ref{tab:responsible-artifacts}). We scanned up to 500,000 documents, skipped documents with empty text or fewer than 50 characters, and tokenized each remaining document without special tokens, truncating at 2,048 tokens. A token was counted only when it immediately followed the single-token article form \texttt{ a} or \texttt{ an} and its decoded string began with a space; tokens without this initial space were excluded as continuation pieces.

\paragraph{Token filters.}
We retained a post-article token type only if it occurred at least 30 times in total, at least 95\% of those occurrences followed one article form, and its decoded string began with a lowercase alphabetic character after stripping whitespace. Usage labels were assigned by the same 95\% criterion: tokens occurring after \emph{an} at that threshold were labeled vowel-initial, and tokens occurring after \emph{a} at that threshold were labeled consonant-initial. Orthographic labels used the first stripped character, with \emph{a, e, i, o, u} treated as vowel letters.

\paragraph{Direction fitting.}
Using the split in Table~\ref{tab:rq1-split}, we fit binary logistic regression to the corresponding rows of the model's \texttt{lm\_head} matrix. The classifier used L-BFGS optimization, $C=1.0$, and a maximum of 1,000 iterations. The normalized coefficient vector defined $\mathbf{d}_1$, with positive scores corresponding to the vowel-initial class.

\paragraph{Stability of the direction.}
Because the classifier is fit with L-BFGS, $\mathbf{d}_1$ does not depend on a random seed, but it can still depend on which trigger tokens happen to enter the training set and on how the cross-validation folds are drawn. We therefore vary both. We refit the direction on 200 bootstrap resamples of the training tokens --- each a training set of the same size, drawn with replacement within each class --- and recompute cross-validation accuracy over 50 redrawn five-fold splits. Both are stable in all three models (Table~\ref{tab:rq1-robustness}): every refit direction points almost the same way as the reported $\mathbf{d}_1$, with a mean cosine similarity of at least 0.989, and cross-validation accuracy varies by at most $10^{-4}$.

\paragraph{Residual directions.}
For the residual-direction analysis, we loaded the same train/test split and projected $\mathbf{d}_1$ out of every row of the \texttt{lm\_head} matrix before fitting a new classifier. Residual classifiers used the same optimization settings as above, with \texttt{class\_weight="balanced"}. After each residual direction was learned, it was projected out before the next iteration. The search was allowed to continue for up to 20 residual directions, but stopped once five-fold cross-validation accuracy fell below 60\%; all three models met this stopping criterion at the first residual direction.

\paragraph{Chance level for the residual classifier.}
Because the training tokens are class-imbalanced (vowel- to consonant-initial $\approx 1{:}4$), 50\% is not the chance level for the residual classifier. We estimated it directly with a label-permutation test, refitting the residual classifier on 200 random shuffles of the training labels. The null cross-validation accuracy is $61.2{\pm}0.6\%$ for Llama, $59.4{\pm}0.6\%$ for Qwen, and $56.9{\pm}0.6\%$ for Gemma. Every residual accuracy in Table~\ref{tab:rq1-iterative} falls below the entire null distribution, including Gemma's 51.74\%.

\paragraph{Non-linear probes.}
To check that this collapse is not an artifact of restricting the probe to a linear decision boundary, we repeated the residual analysis with a multi-layer perceptron,
\[
\hat{y}=\sigma\!\left(\mathbf{w}_2^{\top}\,\mathrm{ReLU}(\mathbf{W}_1\tilde{\mathbf{e}}+\mathbf{b}_1)+b_2\right),
\]
where $\tilde{\mathbf{e}}$ is a standardized embedding row and $\mathbf{W}_1\in\mathbb{R}^{128\times d}$, trained with Adam for at most 300 iterations with early stopping. We report balanced accuracy on the exception set, which is itself imbalanced (about 71\% consonant-initial), so that raw accuracy there is not dominated by the base rate. On the full embeddings the probe separates the exception tokens almost perfectly, but once $\mathbf{d}_1$ is projected out it drops to near chance (Table~\ref{tab:rq1-robustness}), as the linear classifier does.

\begin{table}[t]
\centering
\small
\begin{tabular}{@{}lcccc@{}}
\toprule
\multirow{2}{*}{Model} & \multicolumn{2}{c}{Stability of $\mathbf{d}_1$} & \multicolumn{2}{c@{}}{MLP bal. acc.} \\
\cmidrule(lr){2-3}\cmidrule(l){4-5}
 & Bootstrap cos. & CV-shuffle acc. & Full & Resid. \\
\midrule
Llama & $0.989{\pm}0.001$ & $0.9993{\pm}0.0001$ & 1.000 & 0.542 \\
Qwen & $0.990{\pm}0.0004$ & $0.9990{\pm}0.0001$ & 0.983 & 0.542 \\
Gemma & $0.992{\pm}0.0004$ & $0.9993{\pm}0.0001$ & 1.000 & 0.486 \\
\bottomrule
\end{tabular}
\caption{Robustness checks for the phonological direction. Bootstrap cosine is the cosine similarity between the reported $\mathbf{d}_1$ and the directions refit on 200 resampled training sets; CV-shuffle accuracy is five-fold cross-validation accuracy over 50 redrawn splits. The last two columns give the balanced accuracy of the MLP probe on the exception set, before and after $\mathbf{d}_1$ is projected out. All entries are mean${\pm}$standard deviation where a distribution is available.}
\label{tab:rq1-robustness}
\end{table}

\section{Experimental Details for Section~\ref{sec:rq2}}
\label{app:rq2-details}

\paragraph{Nonce validation.}
To check whether the sampling procedure produced effective nonce bases, we measured closeness to the trigger-token distribution and similarity to existing vocabulary items. Let $E_i$ be the embedding row for token $i$, let $T$ be the retained trigger-token ids, and let $\mu_T=|T|^{-1}\sum_{i\in T}E_i$. For a query vector $x$, centroid distance is
\[
\Delta(x)=\|x-\mu_T\|_2.
\]
Nearest-vocabulary cosine is
\[
N(x)=\max_{k\in V}\frac{x^\top E_k}{\|x\|_2\|E_k\|_2},
\]
with $k=i$ excluded when $x=E_i$ is itself a real trigger-token embedding. Table~\ref{tab:rq2-nonce-validation} reports mean and standard deviation over real trigger tokens and over sampled nonce bases. The nonce bases match real trigger tokens in centroid distance, but have much lower nearest-vocabulary cosine, indicating that they remain close to the trigger-token distribution without collapsing onto existing lexical items.

\begin{table}[t]
\centering
\small
\begin{tabular}{@{}llcc@{}}
\toprule
Model & Set & Centroid dist. & Nearest cos. \\
\midrule
Llama3.2 & Words & $1.111{\pm}0.076$ & $0.684{\pm}0.126$ \\
 & Nonce & $1.114{\pm}0.026$ & $0.226{\pm}0.076$ \\
Qwen2.5 & Words & $1.068{\pm}0.072$ & $0.686{\pm}0.140$ \\
 & Nonce & $1.069{\pm}0.023$ & $0.169{\pm}0.043$ \\
Gemma3 & Words & $0.919{\pm}0.038$ & $0.717{\pm}0.129$ \\
 & Nonce & $0.918{\pm}0.028$ & $0.242{\pm}0.038$ \\
\bottomrule
\end{tabular}
\caption{Validation of the sampled nonce bases. Centroid distance is measured from the mean of the retained trigger-token embeddings. Nearest-vocabulary cosine is the maximum cosine similarity to any vocabulary embedding row; for real trigger tokens, the token's own row is excluded.}
\label{tab:rq2-nonce-validation}
\end{table}

\begin{figure}[t]
\centering
\includegraphics[width=\columnwidth]{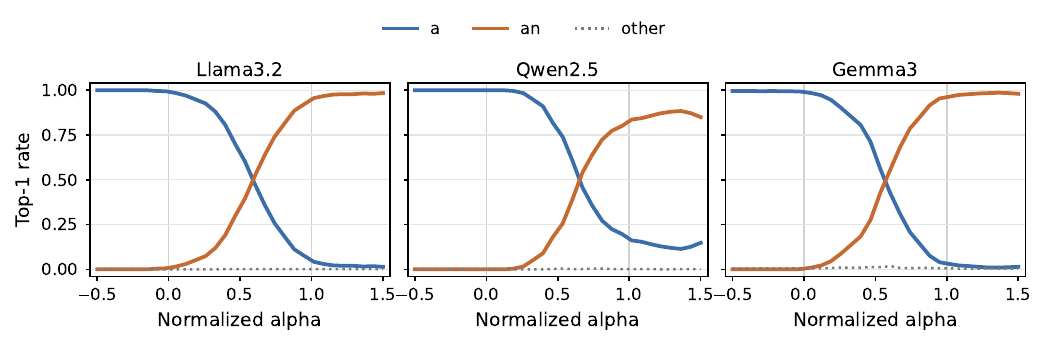}
\caption{Top-1 next-token choices in the token-level wug test. Rates are computed over 500 nonce bases in the Chinese-to-English translation template. The $x$-axis normalizes $\alpha$ so that 0 is the consonant-start class mean and 1 is the vowel-start class mean.}
\label{fig:rq2-wug-top1}
\end{figure}

\paragraph{Top-1 choices.}
The same sweep also produced a categorical shift in the model's top-1 next-token choices (Figure~\ref{fig:rq2-wug-top1}). Near the consonant-start anchor, almost all nonce bases selected \emph{a}; near the vowel-start anchor, most selected \emph{an}, while other top-1 tokens remained rare.

\section{Experimental Details for Section~\ref{sec:rq3}}
\label{app:rq3-details}

\paragraph{Prediction-conditioned contexts.}
Examples were constructed from the same English Wikipedia snapshot \texttt{20231101.en}. Each example consists of the sentence containing the target token preceded by the immediately previous sentence. Table~\ref{tab:rq3-context-examples} shows representative examples from the Llama split. In the \emph{a}/\emph{an} condition, the target token is the article; in the random-token control, the target token is a randomly selected word-like token from the same type of two-sentence context. In both cases, the FutureLens source is the hidden state immediately before the target token, and the future target is the final hidden state at the target-token position, whose decoding predicts the following token.

\begin{table*}[t]
\centering
\scriptsize
\setlength{\tabcolsep}{3pt}
\begin{tabular}{@{}p{0.12\textwidth}p{0.62\textwidth}p{0.09\textwidth}p{0.09\textwidth}@{}}
\toprule
Condition & Context excerpt & Target & Next \\
\midrule
\emph{a}/\emph{an} &
\ldots Natasha Irons / Steel appears in Multiversity as \textbf{a} \underline{member} of the Justice League \ldots &
\emph{a} & member \\
Random &
\ldots explained the properties of the nucleus in terms of these sub-\textbf{atomic} \underline{particles} and the forces \ldots &
atomic & particles \\
\bottomrule
\end{tabular}
\caption{Representative context examples. Bold marks the target token, and underlining marks the following token predicted from the target-token position.}
\label{tab:rq3-context-examples}
\end{table*}

\paragraph{Strict splits.}
We retained only examples in which the target token was the model's top-1 prediction at the preceding position, and the following token was the model's top-1 prediction at the target position. Following tokens were required to contain at least one alphanumeric character. For each model and each condition, the final strict split contains 90,000 training examples and 10,000 held-out test examples; the \emph{a}/\emph{an} split is balanced before splitting, with 50,000 examples per article form. Splits were performed at the text level, so examples from the same two-sentence context do not appear in both train and test sets.

\paragraph{Orthogonal FutureLens.}
The inverse-mapping analysis uses the transpose of the FutureLens matrix, so the final probes constrain the linear map to be orthogonal:
\[
W_\ell^\top W_\ell = I.
\]
We initialized $W_\ell$ with the centered orthogonal Procrustes solution. During optimization, the weight matrix was periodically projected back to the closest orthogonal matrix in Frobenius norm: for a candidate matrix $A=USR^\top$, we set $W_\ell=UR^\top$.

\begin{figure*}[t]
\centering
\includegraphics[width=0.9\textwidth]{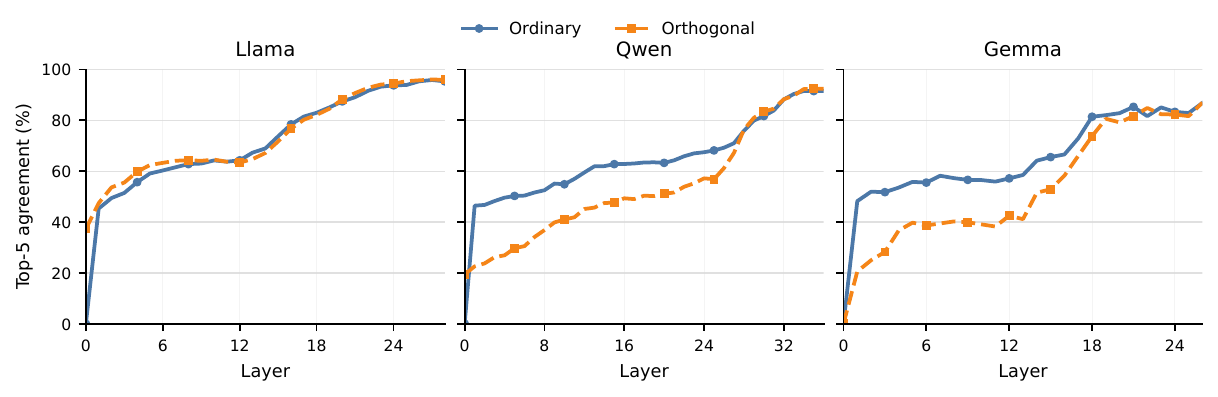}
\caption{Top-5 trigger agreement for ordinary and orthogonal FutureLens maps on matched strict \emph{a}/\emph{an} runs. The orthogonal maps preserve the late-layer forecasting accuracy relevant to the inverse-mapping analysis.}
\label{fig:rq3-orthogonal-comparison}
\end{figure*}

As a check on the constraint itself, Figure~\ref{fig:rq3-orthogonal-comparison} compares ordinary FutureLens maps with orthogonal FutureLens maps on matched strict \emph{a}/\emph{an} runs for Llama, Qwen, and Gemma. The comparison is not intended to establish layerwise equivalence; rather, it shows that in the late layers where trigger forecasting is strongest, the orthogonal maps retain the same peak-level top-5 agreement as ordinary FutureLens maps. Thus, the orthogonality constraint does not explain away the forecasting effect used in the mechanistic analysis.

\begin{figure*}[t]
\centering
\includegraphics[width=0.92\textwidth]{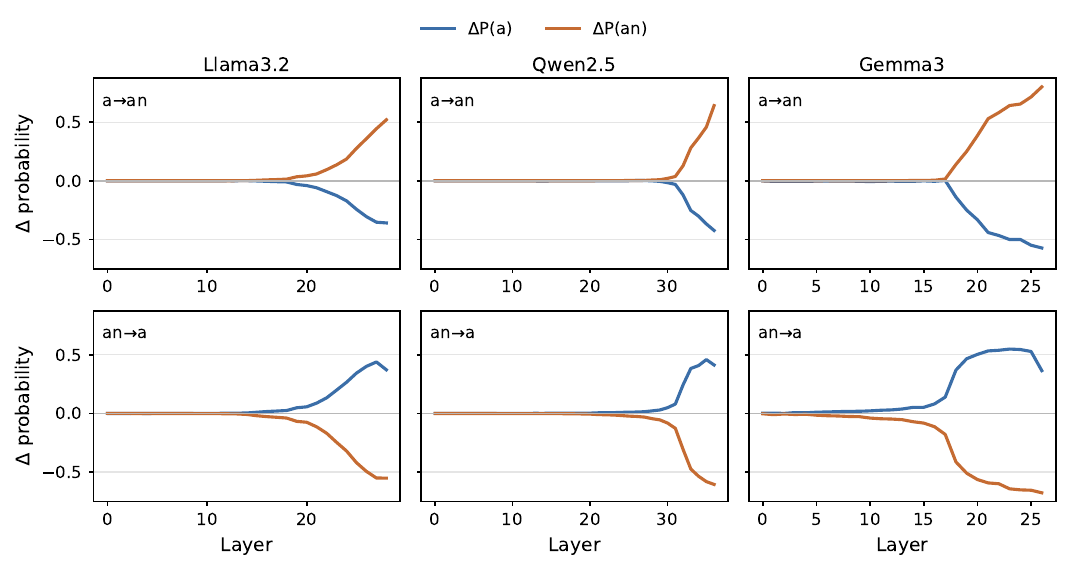}
\caption{Article-probability shifts under counterfactual steering. The top row steers baseline-\emph{a} examples toward the \emph{an} class mean; the bottom row steers baseline-\emph{an} examples toward the \emph{a} class mean. Values are mean probability after intervention minus mean probability before intervention, over up to 1,000 held-out examples per direction.}
\label{fig:rq3-steering-probability}
\end{figure*}

\paragraph{Steering probability shifts.}
The counterfactual steering analysis also records the change in article probabilities, not only the top-1 flip rate reported in Figure~\ref{fig:rq3-steering}. Figure~\ref{fig:rq3-steering-probability} shows that steering baseline-\emph{a} examples toward the \emph{an} class mean increases $P(an)$ and decreases $P(a)$ in the same late layers where flips are most frequent; steering baseline-\emph{an} examples toward the \emph{a} class mean produces the reverse pattern.

\section{The Forecasting Mechanism in a 70B Model}
\label{app:70b}

\paragraph{Motivation.}
The mechanism described in the main text was established in relatively small models of one to three billion parameters, all of which tie their input and output embeddings, and it is not clear whether it also holds in a larger model without weight tying. We therefore apply the experiments of Section~\ref{sec:rq3} to Llama-3.3-70B-Instruct \citep{grattafiori2024llama3herdmodels}, asking whether a model at this scale likewise forecasts the upcoming trigger token at the article-prediction position and uses the forecasted trigger's phonological direction to choose between \emph{a} and \emph{an}.

\paragraph{Setup.}
The model has 80 layers and a hidden size of 8,192. The corpus, the strict filtering criteria, the probe objective, the orthogonality constraint, and the steering protocol are unchanged. The strict \emph{a}/\emph{an} split contains 80,000 training and 20,000 held-out test examples, again balanced with 50,000 examples per article form, and the random-token control is built the same way. Because this model does not tie its embeddings, we obtain $\mathbf{d}$ in the unembedding space by fitting the classifier of Section~\ref{sec:rq1} to the rows of $W_{\text{out}}$ instead of $E$. To limit compute, we fit probes at every second layer from layer 40 upward, the upper half of the network, where both effects are strongest in the smaller models.

\begin{figure}[t]
\centering
\includegraphics[width=\columnwidth]{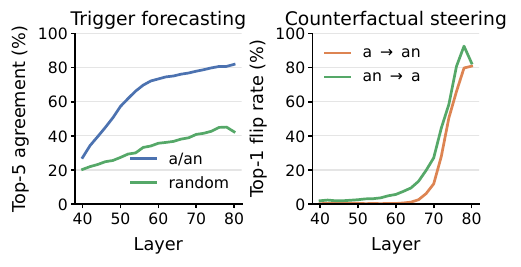}
\caption{The forecasting mechanism in Llama-3.3-70B-Instruct. Left: FutureLens top-5 agreement with the ground-truth trigger token at $t+2$, for the \emph{a}/\emph{an} condition and the random-token control. Right: top-1 flip rate under counterfactual steering along $\mathbf{s}_\ell$, for baseline-\emph{a} and baseline-\emph{an} examples. Probes are fit at every second layer from 40 to 80.}
\label{fig:rq3-70b}
\end{figure}

\paragraph{Results.}
Both effects reappear at this scale (Figure~\ref{fig:rq3-70b}). FutureLens top-5 agreement in the \emph{a}/\emph{an} condition peaks at 81.8\% in the final layer, against 45.1\% for the random-token control, a separation comparable to the one observed in the smaller models. Counterfactual steering along $\mathbf{s}_\ell$ flips the top-1 article for an average of 86.1\% of held-out examples at layer 78 (79.7\% from \emph{a} to \emph{an} and 92.5\% from \emph{an} to \emph{a}). As in the smaller models, the steering effect is confined to the final layers and emerges later in depth than the forecast itself.

The same account therefore holds in a model more than twenty times larger than any in our main analysis: the article-prediction position carries a forecast of the upcoming trigger token, and a single phonological direction carried by that forecast has a strong causal effect on the article decision. Because that direction is learned here in the unembedding space and mapped back through $W_\ell^\top$, the result also shows that the inverse FutureLens intervention does not require weight tying.

\section{Experimental Details for Section~\ref{sec:rq4}}
\label{app:rq4-details}

\paragraph{Cross-linguistic templates.}
Table~\ref{tab:rq4-templates} summarizes the allomorphy systems, examples, translation prompts, assistant prefixes, and scored allomorph families.

\begin{table*}[t]
\centering
\scriptsize
\setlength{\tabcolsep}{5pt}
\renewcommand{\arraystretch}{1.12}
\begin{tabular}{@{}p{0.13\textwidth}p{0.48\textwidth}p{0.31\textwidth}@{}}
\toprule
\multicolumn{3}{@{}l}{\textbf{A. Allomorphy systems}} \\
\midrule
Language & Allomorphy & Examples \\
\midrule
French &
Definite article contraction: \emph{le}/\emph{la} before consonant-initial nouns, \emph{l'} before vowel-initial nouns. &
\emph{le livre} `the book'; \emph{la maison} `the house'; \emph{l'école} `the school' \\
\addlinespace
Korean &
Object particle alternation: \hangultext{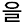} after consonant-final nouns, \hangultext{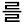} after vowel-final nouns. &
\hangultext{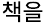} `book-OBJ'; \hangultext{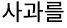} `apple-OBJ' \\
\addlinespace
Turkish &
Locative vowel harmony: front-vowel \emph{-de}/\emph{-te} vs. back-vowel \emph{-da}/\emph{-ta}. &
\emph{evde} `at home'; \emph{okulda} `at school'; \emph{parkta} `at the park' \\
\addlinespace
Italian &
Masculine indefinite article alternation between \emph{un} and \emph{uno}. &
\emph{un libro} `a book'; \emph{uno studente} `a student' \\
\midrule
\multicolumn{3}{@{}l}{\textbf{B. Evaluation templates}} \\
\midrule
Language & User prompt and assistant prefix & Scored families \\
\midrule
French &
\texttt{Translate to French.}\newline
\texttt{I saw the <nonce>.}\newline
Prefix: \emph{J'ai vu} &
\emph{le}/\emph{la} vs. \emph{l'} \\
\addlinespace
Korean &
\texttt{Translate to Korean: I saw<nonce>.}\newline
Prefix: \hangultext{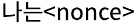} &
\hangultext{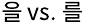} \\
\addlinespace
Turkish &
\texttt{Translate to Turkish.}\newline
\texttt{I work at the <nonce> every day.}\newline
Prefix: \emph{Her gün <nonce>} &
\emph{de}/\emph{te} vs. \emph{da}/\emph{ta} \\
\addlinespace
Italian &
\texttt{Translate to Italian.}\newline
\texttt{I saw a<nonce>.}\newline
Prefix: \emph{Ho visto} &
\emph{un} vs. \emph{uno} \\
\bottomrule
\end{tabular}
\caption{Cross-linguistic allomorphy examples and translation templates. The \texttt{<nonce>} placeholder denotes the nonce token whose embedding is replaced during the sweep. Korean forms are shown in Hangul.}
\label{tab:rq4-templates}
\end{table*}

\paragraph{Cross-linguistic scoring.}
Direction-learning counts were collected from the Hugging Face \texttt{wikimedia/wikipedia} dataset, using the \texttt{20231101.fr}, \texttt{20231101.ko}, \texttt{20231101.tr}, and \texttt{20231101.it} splits. For each language, we streamed up to 100,000 documents, skipped empty documents and documents shorter than 50 characters, tokenized each remaining document without special tokens, and truncated it to 2,048 model tokens.

Token types were labeled by the allomorphic context in which they occurred. French counted word-start lexical tokens following \emph{le}/\emph{la} against raw non-word-start lexical tokens following \emph{l'}; Korean counted word-start tokens immediately preceding \hangultext{figures/rq4_ko_eul} against those preceding \hangultext{figures/rq4_ko_reul}; Turkish counted tokens preceding front-vowel locative suffix sequences (\emph{de}, \emph{te}, \emph{'de}, \emph{'te}) against those preceding back-vowel sequences (\emph{da}, \emph{ta}, \emph{'da}, \emph{'ta}), requiring the suffix to end at a word boundary and excluding leading-space suffix tokens; Italian counted word-start lexical tokens following \emph{un} against those following \emph{uno}, excluding empty, punctuation-only, nonalphabetic, and uppercase tokens. A token type was retained only if at least 95\% of its counted occurrences fell on one side of the contrast and it met the selected minimum count: 30 for French and Korean, 5 for Turkish, and 15 for Italian.

After fitting the language-specific direction, we sampled nonce embeddings from the empirical distribution of the retained token embeddings, removed their component along the learned direction, and swept that component using the same midpoint-centered rule as in the English wug test. At each sweep point, we scored the allomorph family immediately after the assistant prefix. When an allomorph family can be represented by more than one tokenizer sequence, we sum the sequence probabilities for that family before computing the plotted rates.

\paragraph{Direct sound-judgment setup.}
The direct sound-judgment experiment uses the same 500 nonce embeddings and the same midpoint-centered $\alpha$ sweep as the English article-choice wug test. The prompt has two variants, asking whether the nonce noun in the phrase \texttt{the<nonce>} begins with a consonant or vowel sound, with no intervening space after \emph{the}. The assistant prefix is \texttt{Answer:}, so the measurement targets the model's next-token answer to the yes/no question.

\paragraph{Sound-judgment scoring.}
For each prompt variant and sweep value, we score the next token after \texttt{Answer:}. The \emph{yes} family sums single-token variants with and without an initial space and with lowercase or capitalized spelling; the \emph{no} family is defined analogously. The experiment records mean $P(yes)$ and $P(no)$ over nonce tokens, and also records whether the top-1 next token falls in the \emph{yes} family, the \emph{no} family, or neither. Figure~\ref{fig:rq4-sound-judgment} plots $P(yes)-P(no)$, the direct contrast for the binary judgment.

\section{The Forecasting Mechanism in French and Italian}
\label{app:rq4-futurelens}

\paragraph{Motivation.}
Section~\ref{sec:rq4} extended the analysis to other languages only at the level of the trigger token's embedding. Here we ask whether the mechanism described in Section~\ref{sec:rq3} also operates in models of other languages. We use French and Italian, the two languages in our set in which the trigger follows the allomorph, so that the model must commit to a form before the trigger token is available. In Korean and Turkish the trigger precedes the allomorph, and there is no upcoming trigger to forecast.

\paragraph{Setup.}
We repeat the experiments of Section~\ref{sec:rq3} on the French and Italian models of Table~\ref{tab:rq4-languages}, using the Wikipedia snapshots and the token labels described in Appendix~\ref{app:rq4-details}. The strict filtering criteria, the probe objective, the orthogonality constraint, and the steering protocol are unchanged. Each strict split contains 80,000 training and 20,000 held-out test examples, balanced with 50,000 examples per allomorph class before splitting, and the random-token control is built the same way. Neither model ties its embeddings, so, as in Appendix~\ref{app:70b}, the language-specific direction $\mathbf{d}$ is obtained in the unembedding space, from the rows of $W_{\text{out}}$ rather than $E$. French needs one adjustment: Lucie tokenizes \emph{l'} as two tokens, while \emph{le} and \emph{la} are single tokens, so we align every example on the state immediately before the lexical trigger, pool the two classes during training, and report evaluation separately for each class.

\begin{figure}[t]
\centering
\includegraphics[width=\columnwidth]{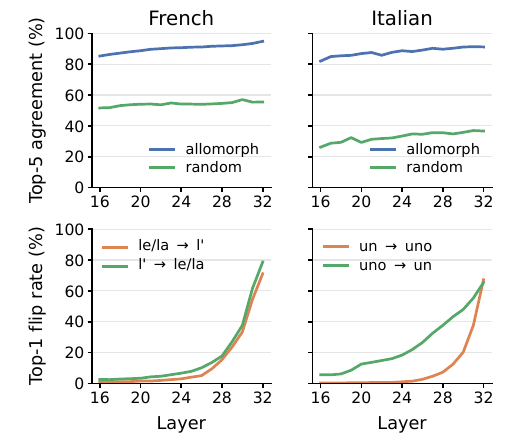}
\caption{The forecasting mechanism in French and Italian. Top: FutureLens top-5 agreement with the ground-truth trigger token, for the allomorph condition and the random-token control. Bottom: top-1 flip rate under counterfactual steering along $\mathbf{s}_\ell$, in both directions. Probes are fit at every layer from 16 to 32.}
\label{fig:rq4-futurelens}
\end{figure}

\paragraph{Results.}
Both effects reappear in the two languages (Figure~\ref{fig:rq4-futurelens}). FutureLens top-5 agreement with the ground-truth trigger token peaks at 94.8\% in the final layer for French, against 57.0\% for the random-token control, and at 91.4\% at layer 31 for Italian, against 36.9\% for its control. Counterfactual steering along $\mathbf{s}_\ell$ flips the top-1 allomorph in the final layer for an average of 74.9\% of held-out French examples (71.1\% from \emph{le}/\emph{la} to \emph{l'} and 78.7\% in the reverse direction) and 66.3\% of Italian examples (67.2\% from \emph{un} to \emph{uno} and 65.4\% in the reverse direction). As in English, the forecast is already accurate in the middle of the network, while the steering effect grows sharply only in the last few layers.

The allomorph-prediction position in these two languages therefore behaves as it does in English: it carries a forecast of the upcoming trigger token, and a single direction within that forecast has a strong causal effect on the allomorph decision. We take this as evidence that a similar mechanism supports allomorph selection beyond English \emph{a}/\emph{an}.

\section{Responsible Research and Reproducibility Details}
\label{app:responsible-research}

\paragraph{Artifacts and terms of use.}
Table~\ref{tab:responsible-artifacts} lists the main external artifacts used in this study. We used the pretrained models only for offline research analysis and did not fine-tune, redistribute, or deploy them. All model access and use followed the terms listed on the corresponding Hugging Face model cards at the time of the experiments. The derived artifacts produced by this work, including learned directions, nonce embeddings, probes, and aggregate results, are intended for research and reproducibility rather than for deployment as model modifications.

\begin{table*}[t]
\centering
\scriptsize
\setlength{\tabcolsep}{3pt}
\begin{tabular}{@{}p{0.22\textwidth}p{0.18\textwidth}p{0.12\textwidth}p{0.18\textwidth}p{0.22\textwidth}@{}}
\toprule
Artifact & Creator/source & Size & License or terms & URL \\
\midrule
Llama-3.2-3B-Instruct \citep{grattafiori2024llama3herdmodels} & Meta & 3B & Llama 3.2 Community License & \url{https://huggingface.co/meta-llama/Llama-3.2-3B-Instruct} \\
Llama-3.3-70B-Instruct \citep{grattafiori2024llama3herdmodels} & Meta & 70B & Llama 3.3 Community License & \url{https://huggingface.co/meta-llama/Llama-3.3-70B-Instruct} \\
Qwen2.5-3B-Instruct \citep{qwen2.5} & Alibaba Cloud/Qwen Team & 3B & Qwen Research License Agreement & \url{https://huggingface.co/Qwen/Qwen2.5-3B-Instruct} \\
Gemma-3-1b-it \citep{gemma_2025} & Google & 1B & Gemma Terms of Use & \url{https://huggingface.co/google/gemma-3-1b-it} \\
EXAONE-4.0-1.2B \citep{exaone-4.0} & LG AI Research & 1.2B & EXAONE AI Model License Agreement 1.2 -- NC & \url{https://huggingface.co/LGAI-EXAONE/EXAONE-4.0-1.2B} \\
Trendyol-LLM-8B-T1 \citep{trendyolLLM8BT1} & Trendyol & 8B & Apache-2.0 & \url{https://huggingface.co/Trendyol/Trendyol-LLM-8B-T1} \\
Minerva-7B-instruct-v1.0 \citep{orlando-etal-2024-minerva} & Sapienza NLP & 7B & Apache-2.0 & \url{https://huggingface.co/sapienzanlp/Minerva-7B-instruct-v1.0} \\
Lucie-7B-Instruct-v1.1 \citep{openllm2025lucie} & OpenLLM-France & 7B & Apache-2.0 & \url{https://huggingface.co/OpenLLM-France/Lucie-7B-Instruct-v1.1} \\
Wikipedia snapshots \texttt{20231101.en}, \texttt{20231101.fr}, \texttt{20231101.ko}, \texttt{20231101.tr}, \texttt{20231101.it} & Wikimedia via Hugging Face Datasets & -- & CC BY-SA 3.0 and GFDL & \url{https://huggingface.co/datasets/wikimedia/wikipedia} \\
\bottomrule
\end{tabular}
\caption{External artifacts used in the experiments. License and terms are reported as listed on the linked model or dataset cards.}
\label{tab:responsible-artifacts}
\end{table*}

\paragraph{Data privacy and content.}
Our corpus-derived examples come from public Wikipedia text rather than from newly collected private or user-level data. Wikipedia may contain names of public or private individuals and may contain offensive material in some articles. We did not collect user metadata, infer demographic attributes, or attempt to identify individuals. The analysis reduces the corpus to token counts, filtered token types, aggregate statistics, and model-internal representations; we do not redistribute the raw Wikipedia corpus, and the paper includes only short illustrative excerpts.

\paragraph{Compute and software.}
The experiments were run on a single NVIDIA DGX Spark system and two NVIDIA A100 GPUs, and took approximately 96 GPU-hours in total. The software environment used the NVIDIA PyTorch Docker image \texttt{nvcr.io/nvidia/pytorch:25.12-py3}, with PyTorch, Hugging Face \texttt{transformers}, \texttt{datasets}, and \texttt{accelerate}, \texttt{scikit-learn}, \texttt{nltk}, and \texttt{matplotlib}. Experiment-specific data sizes, splits, hyperparameters, and aggregation procedures are reported in Appendices~\ref{app:rq1-details}--\ref{app:rq4-futurelens}.

\paragraph{AI assistance.}
We used Codex and Claude Code for experimental coding and writing edits. All outputs were manually reviewed by the authors before incorporation.

\end{document}